\documentclass[10pt,twocolumn]{ICCAS}
 
\usepackage{diagbox}
\usepackage{multirow}
\usepackage{subcaption}
\usepackage{url}  

\begin{document}

\title{Self Supervised Learning from Automatically Generated Demonstrations for Visual Robotic Manipulation}

\author{Andres Rivas${}^{1}$, Anselmo R. Cukla${}^{2}$, Rodrigo S. Guerra${}^{3}$, Bruna V. Guterres${}^{1*}$ and Ricardo B. Grando${}^{1*}$ }

\affils{ ${}^{1}$Robotics and AI Lab, Technological University of Uruguay, \\
Rivera, Uruguay (ricardo.bedin@utec.edu.uy) \\
${}^{2}$Center of Technology, Federal University of Santa Maria, \\
Santa Maria, Brazil (cukla.anselmo@gmail.com) \\
${}^{3}$Computer Science Center, Federal University of Rio Grande, \\
Rio Grande, Brazil (tioguerra@gmail.com) {\small${}^{*}$ Corresponding author}}


\abstract{
Robotic manipulation often requires object specific programming, manual data annotation, or calibrated perception pipelines, which limits rapid deployment in practical settings. Learning from demonstration offers a more direct alternative, but collecting demonstrations can still demand human teleoperation or kinesthetic teaching \cite{johns2021,wang2023mimicplay,young2020visual}. This paper presents a self supervised visual manipulation method in which a robot automatically generates demonstrations around a target pose and learns relative pose corrections directly from wrist mounted RGB images. The proposed pipeline uses ROS~2 and Isaac Sim to collect labeled image-pose pairs without requiring explicit camera to robot extrinsic calibration. Separate datasets are generated for planar refinement and coarse three dimensional approach, and a convolutional network is trained to regress relative translation and rotation from single frame RGB observations \cite{johns2021}. During execution, a coarse to fine controller first approaches the object using models trained with height variation and then refines the final alignment using planar data. The method is evaluated both in simulation and on a real UR5e collaborative robot equipped with a gripper and a monocular camera. In simulation, the refinement stage reduces the final planar dispersion from 9.69 mm to 5.38 mm. In real world experiments, the system performs end to end grasp attempts on three physical objects and reaches success rates of 66.6\% and 63.6\% for two objects without object rotation, while still maintaining partial robustness under rotated conditions. These results show that automatically generated demonstrations can support practical visual manipulation with limited setup effort, while also exposing remaining challenges in depth prediction and object dependent generalization.
}


\maketitle


\section{Introduction}

Robotic manipulation remains difficult to deploy outside structured industrial environments. Traditional approaches often require manual programming, calibrated perception pipelines, or object-specific modeling, which limits flexibility when new tasks or objects are introduced. These constraints increase engineering effort and reduce the scalability of robotic systems in dynamic or unstructured settings.

\begin{figure*}[t]
    \centering
    \includegraphics[width=0.89\textwidth]{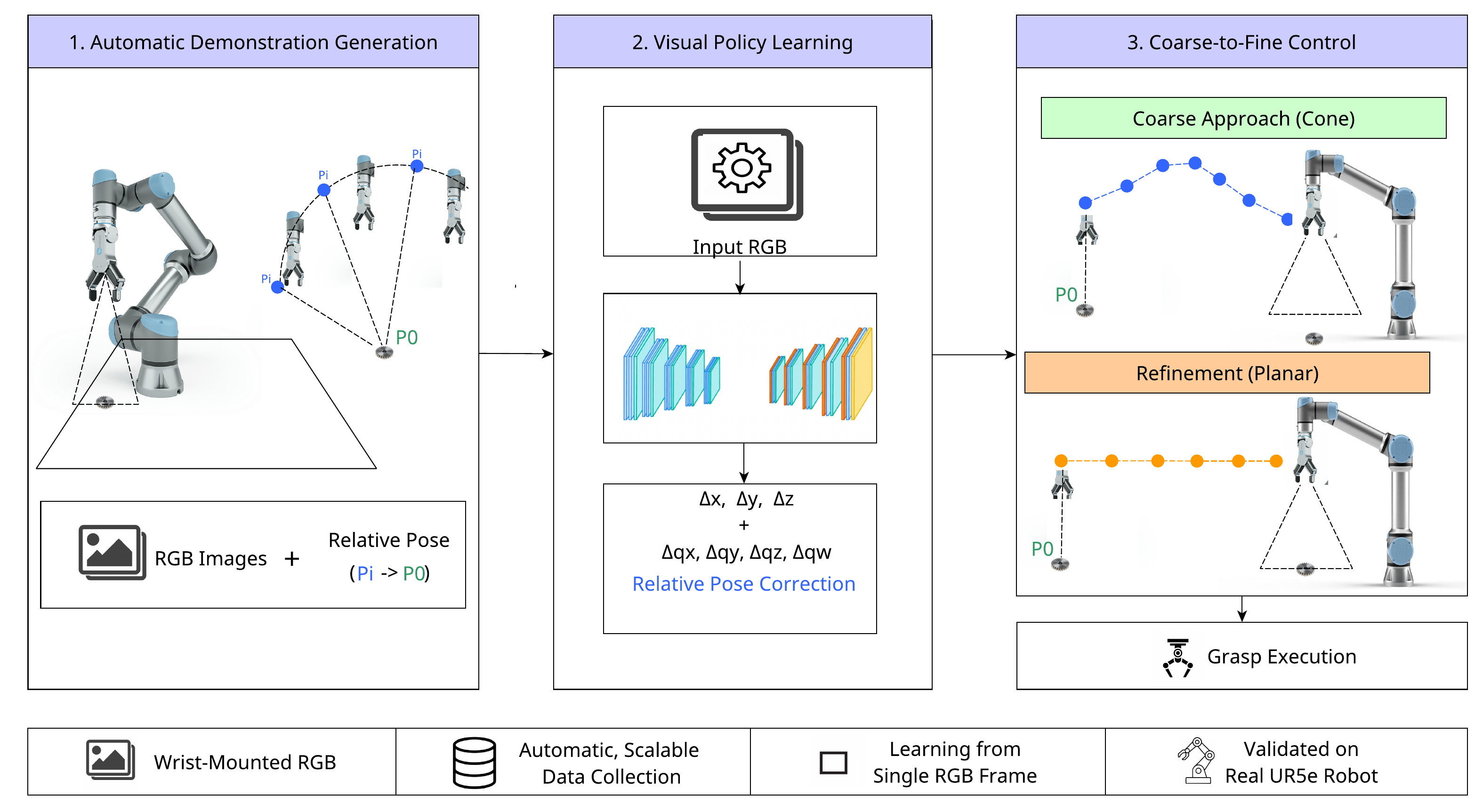}
    \caption{Overview of the proposed framework. A wrist-mounted monocular camera on a UR5e robot is used to automatically generate demonstrations around a reference pose $P_0$. These image-pose pairs are used to train a visual policy that predicts relative pose corrections from RGB observations. During execution, a coarse-to-fine controller combines approach and refinement stages and is validated on a real UR5e platform.}
    \label{fig:paper_overview}
\end{figure*}

Visual learning methods offer an alternative by allowing robot actions to be inferred directly from sensory observations. In particular, learning from demonstration enables robots to acquire manipulation behavior from examples rather than explicit programming. However, most existing approaches rely on human-provided demonstrations, such as kinesthetic teaching or teleoperation, which can be time-consuming and difficult to scale \cite{wang2023mimicplay,young2020visual}. Moreover, many visual manipulation pipelines still depend on explicit camera calibration or structured perception stages, which adds complexity to real-world deployment.

Recent work has explored the possibility of reducing these requirements through synthetic data generation, domain randomization, and automated supervision \cite{tobin2017domain,labbe2022megapose}. In this direction, coarse-to-fine imitation learning has shown that robots can generate demonstrations around a target pose and learn relative corrections from visual input \cite{johns2021}. This idea suggests that demonstration collection itself can be automated, enabling a self-supervised formulation of visual manipulation.

This paper builds on that perspective and proposes a self-supervised framework for visual robotic manipulation based on automatically generated demonstrations. A wrist-mounted monocular camera observes the scene while the robot explores poses around a reference manipulation pose. Each observation is paired with the relative transformation required to return to the target, producing training data without manual annotation or explicit camera-to-robot calibration. A convolutional model is trained to predict relative pose corrections from RGB images, and a coarse-to-fine control strategy is used during execution to combine three-dimensional approach and planar refinement.

An overview of the proposed pipeline is shown in Fig.~\ref{fig:paper_overview}. The system\footnote{Video available at: \url{https://youtu.be/PSGrWwR8Rug?si=8eH41TkTl3W5whPM}} integrates automatic data collection, visual policy learning, and closed-loop control within a unified ROS~2 framework, enabling direct transfer from simulation to real hardware. Unlike prior approaches that focus on low-dimensional or simulated settings, this work extends self-supervised demonstration generation to real-world visual manipulation with object-specific datasets and closed-loop grasp validation.

The main contributions of this paper are:
\begin{itemize}
    \item A self-supervised data collection method for visual robotic manipulation, where the robot automatically generates demonstration-like image--pose pairs without manual labeling or human intervention.
    \item A visual policy that learns relative pose corrections directly from monocular RGB observations, without requiring explicit camera-to-robot extrinsic calibration.
    \item A coarse-to-fine control strategy that separates three-dimensional approach from planar refinement, improving final alignment in closed-loop execution.
    \item A full validation of the proposed framework in both simulation and real-world experiments on a UR5e robot, demonstrating practical manipulation performance across multiple real-world objects.
\end{itemize}

\vspace{-5mm}
\section{Related Work}

\subsection{Data driven pose estimation and tracking}

Several visual pose estimation methods learn object pose directly from images. PoseCNN estimates object translation and rotation from RGB observations through a convolutional architecture \cite{xiang2018posecnn}. DeepIM improves pose estimates through iterative image based refinement \cite{li2019deepim}. se(3)-TrackNet predicts relative pose between current and previous RGB-D observations, showing the value of learned pose tracking under synthetic training conditions \cite{wen2020se3tracknet}. More recent methods such as CosyPose and MegaPose combine synthetic rendering, object models, and staged refinement to estimate pose for known or novel objects \cite{labbe2020cosypose,labbe2022megapose}. These works show that learned vision models can achieve accurate pose estimation, but many still rely on object models, calibrated inputs, or structured data generation.

\subsection{Dataset generation for manipulation and pose estimation}

Benchmark datasets such as YCB and T-LESS have been widely used for pose estimation research, but their collection process usually depends on controlled acquisition hardware, multiple sensors, or manual annotation \cite{calli2015benchmarking,hodan2017tless,hodan2018bop}. Xiang et al. \cite{xiang2018posecnn} reduced the labeling burden by annotating only the first frame of each sequence and recovering the rest through tracking. At the same time, fully synthetic training has become more common because it simplifies annotation and increases variation through randomization \cite{labbe2022megapose,tremblay2018deep,wen2020se3tracknet,tobin2017domain}. Automatic dataset generation pipelines have also been explored for grasping and vision based manipulation, reinforcing the practical value of reducing manual labeling effort \cite{ahmad2021automatic}. This trend is important for manipulation, since automatically generated training data can reduce setup time and improve reproducibility. Still, many synthetic approaches remain focused on pose estimation rather than closed loop manipulation.

\subsection{Learning from demonstration and self supervised demonstration generation}

Imitation learning methods learn robot actions from demonstrations collected by human guidance, teleoperation, or video observation \cite{wang2023mimicplay,young2020visual}. These methods are attractive because they simplify task specification, but their deployment cost depends on how demonstrations are collected and post processed. Johns \cite{johns2021} proposed a coarse to fine imitation learning framework in which the robot automatically generates demonstrations relative to a target pose, thereby reducing manual effort. The present work is closest to that formulation. However, unlike the 3-DOF setup described in \cite{johns2021}, the proposed method extends the idea to object specific visual manipulation with staged datasets of increasing complexity, integrates ROS~2 based control in simulation and on a UR5e robot, and places particular emphasis on real world grasp validation.

\vspace{-6mm}
\section{Methodology}

This section describes the proposed self supervised framework for visual robotic manipulation. The method is organized as a pipeline that starts with the definition of a reference manipulation pose, continues with the automatic generation of image-pose pairs around that reference, and ends with the training and deployment of a visual controller. The same general procedure is used in simulation and on the physical robot, which makes it possible to develop the system offline and later validate it in a real environment. The overall goal is to learn relative pose corrections from RGB observations without requiring explicit camera to robot extrinsic calibration.

\subsection{System Overview}

The proposed system learns a mapping from a single RGB image to the relative pose correction needed to move the robot tool center point from its current pose $P_i$ to a target manipulation pose $P_0$. The full implementation is built on ROS~2, which allows the same communication and control structure to be used across simulation and real hardware.

In the simulated setup, Isaac Sim is used to generate RGB images and robot state information through ROS~2 bridges. A UR5e model, a camera mounted relative to the robot tool, a light source, and 3D object models are included in the scene. The simulated camera publishes images through the RGB topic, while joint states and robot transforms are exchanged between Isaac Sim and the robot-side learning pipeline. In the physical setup, the same learning pipeline is connected to a real UR5e robot equipped with an OnRobot RG6 gripper and a Logitech USB camera mounted on the robot. The camera publishes RGB frames through a ROS~2 camera node, while the robot controller is accessed through the \texttt{ur\_robot\_driver}. 

\begin{figure*}[t]
    \centering
    \begin{subfigure}{0.46\textwidth}
        \centering
        \includegraphics[width=\linewidth]{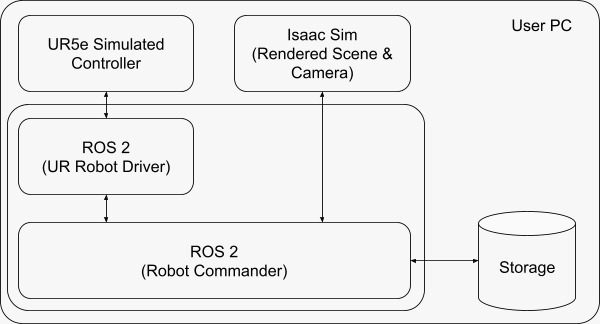}
        \caption{Software components of the simulated setup.}
        \label{fig:sim_sw_components}
    \end{subfigure}
    \hfill
    \begin{subfigure}{0.46\textwidth}
        \centering
        \includegraphics[width=\linewidth]{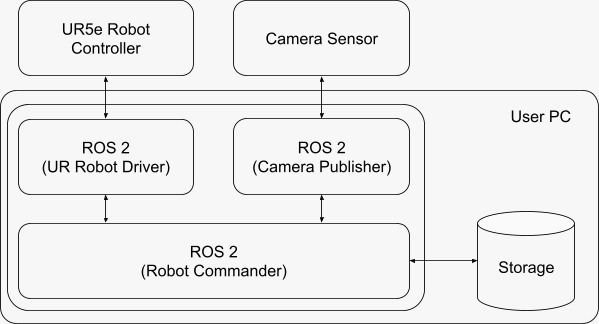}
        \caption{Software components of the real setup.}
        \label{fig:real_sw_components}
    \end{subfigure}
    \caption{Software components of the proposed system in simulation and in the real world. In both cases, the learning pipeline is built on ROS~2. The main difference is that the simulated setup uses URSim and Isaac Sim as data and state sources, while the real setup uses the physical UR5e controller and a USB camera.}
    \label{fig:sw_components}
\end{figure*}

Figure~\ref{fig:sw_components} summarizes the software components in both cases. In simulation, shown in Fig.~\ref{fig:sim_sw_components}, the \texttt{ur\_robot\_driver} communicates with URSim, while Isaac Sim provides rendered RGB observations through ROS~2 to the \textit{Robot Commander} node. In the real setup, shown in Fig.~\ref{fig:real_sw_components}, the same ROS~2 pipeline is preserved, but the rendered image source is replaced by a physical USB camera and the simulated controller is replaced by the real UR5e controller. This shared structure is important because it allows the data generation and learning pipeline to be developed in simulation and then transferred to the real platform with only minimal interface changes.


\subsection{Automatic Generation of Demonstrations}

Training data are generated automatically by a ROS node that commands the robot along predefined trajectories while synchronizing each RGB image with its corresponding relative pose label. A reference pose $P_0$ is first defined with respect to the target object. This pose represents the desired manipulation pose that the robot should reach before executing the grasping motion. During data collection, the robot is moved through a set of poses $P_i$ generated around $P_0$, and for every received image the relative transformation from $P_i$ to $P_0$ is computed and stored as ground truth. In this sense, each sample acts as an automatically generated demonstration, since it pairs a visual observation with the correction that should be executed to reach the target pose \cite{johns2021}.

The data collection procedure is implemented through a ROS node denoted as \textit{Robot Commander}. This node commands the robot motion, subscribes to the RGB topic, and computes the corresponding pose label inside the image callback. The relative transform between the current pose and the reference pose is obtained through \texttt{tf2}, which allows the current camera or tool pose to be related to the reference pose at the exact time each frame is received. As illustrated in Fig.~\ref{fig:real_sw_components}, the same real setup architecture is used during data generation: the robot state is provided through \texttt{ur\_robot\_driver}, the camera images are published by the USB camera node, and the resulting samples are stored for offline training. This removes the need for manual annotation and allows the same collection logic to be reused in simulation and on the real robot.

\begin{figure}[t]
    \centering
    \begin{subfigure}{0.1\textwidth}
        \includegraphics[width=\linewidth]{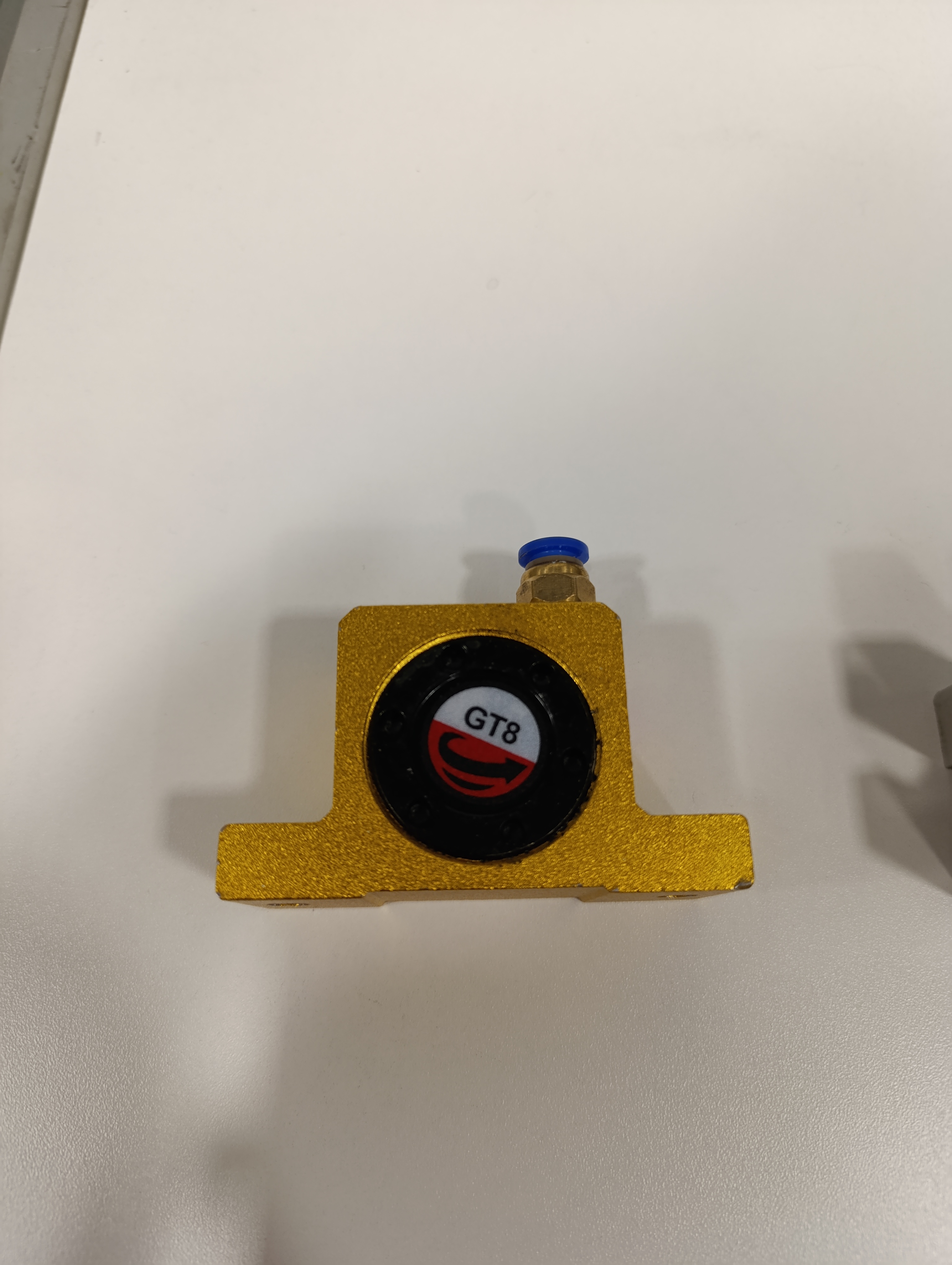}
        \caption{Object 1}
        \label{fig:obj1}
    \end{subfigure}
    \begin{subfigure}{0.1\textwidth}
        \includegraphics[width=\linewidth]{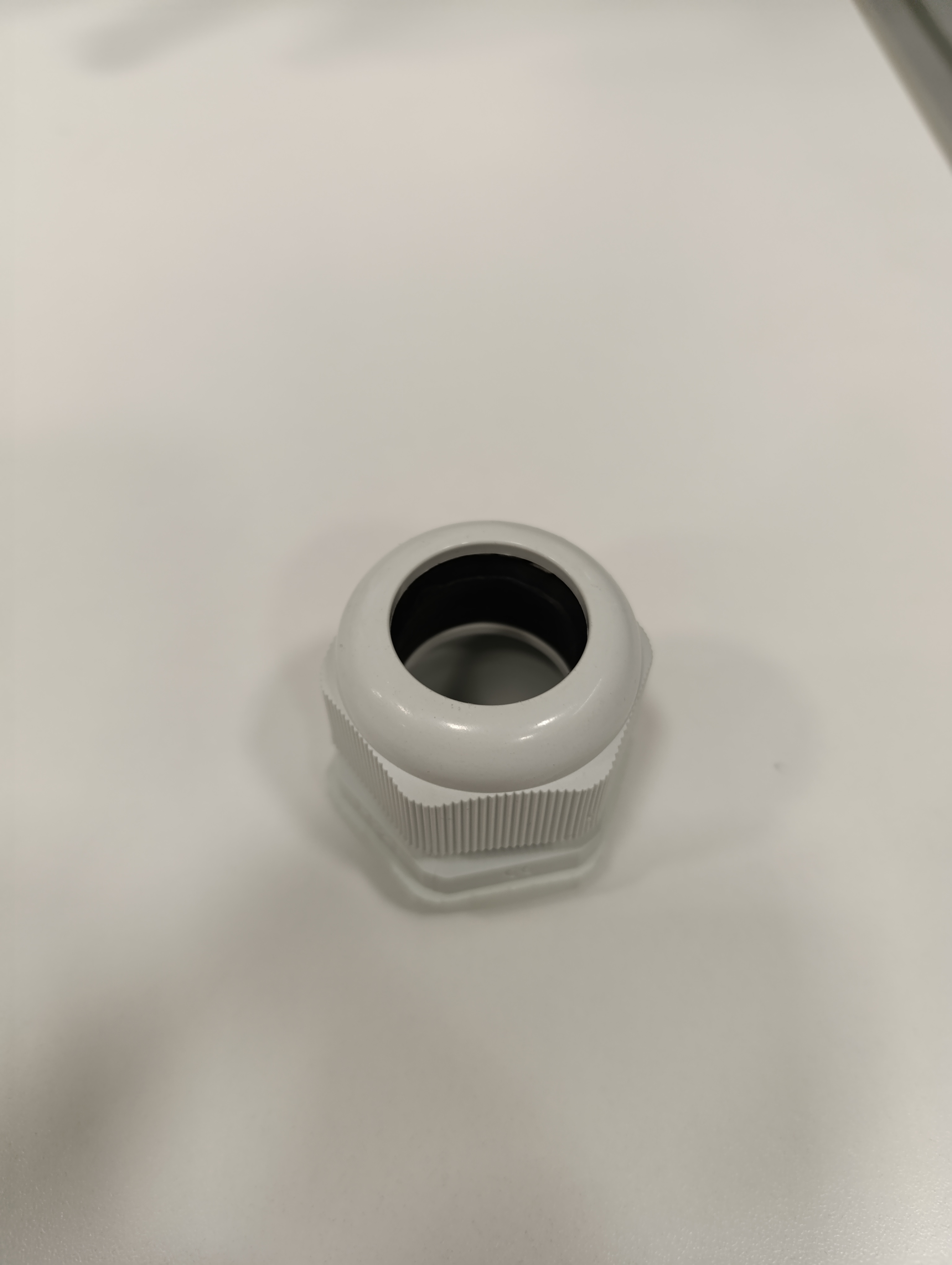}
        \caption{Object 2}
        \label{fig:obj2}
    \end{subfigure}
    \begin{subfigure}{0.1\textwidth}
        \includegraphics[width=\linewidth]{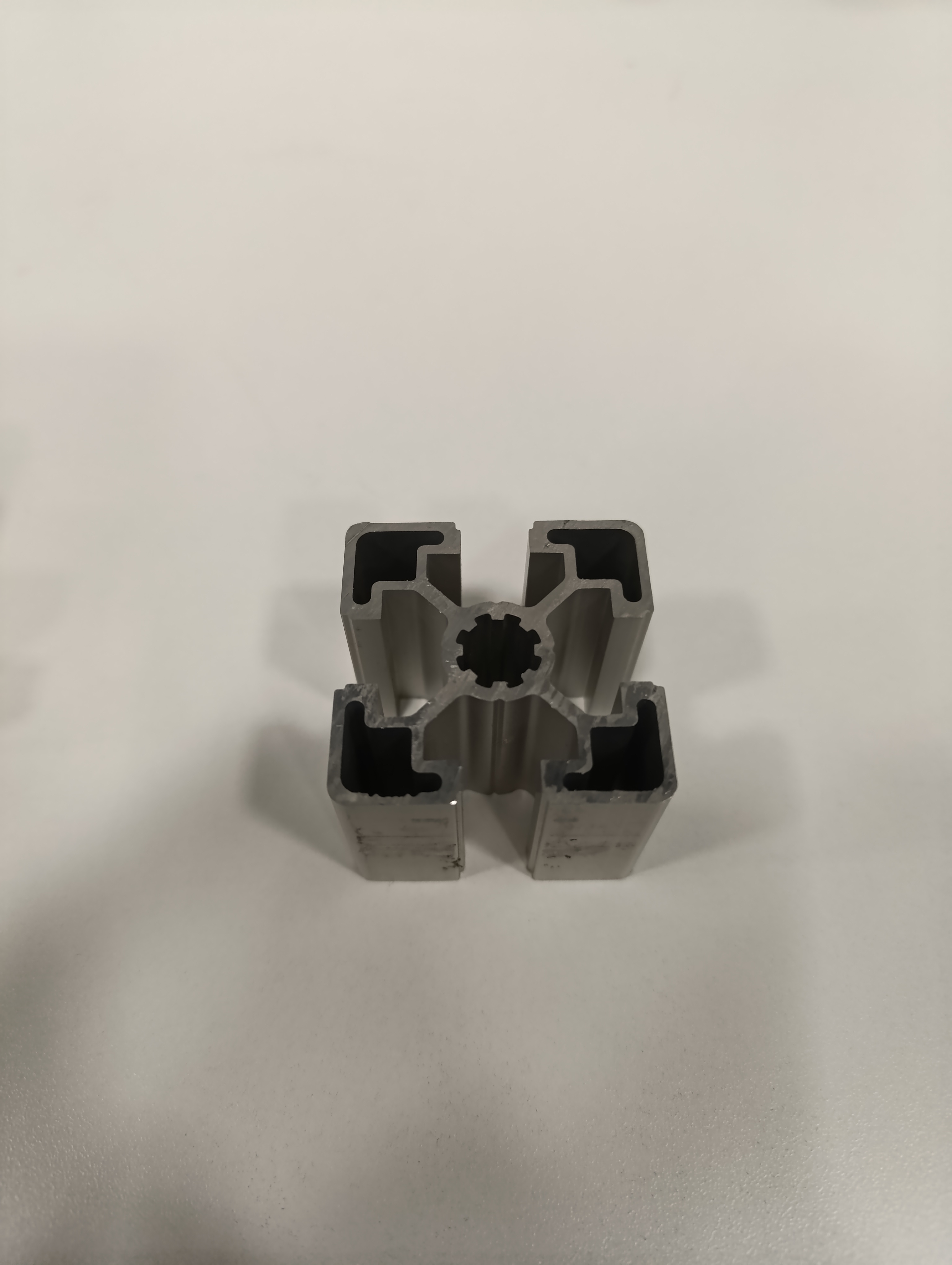}
        \caption{Object 3}
        \label{fig:obj3}
    \end{subfigure}
    \caption{Physical objects used for the real-world experiments and the automatic dataset generation procedure.}
    \label{fig:all-obj}
\end{figure}

A separate dataset was recorded for each of the three physical objects used in the experiments. To preserve anonymization in the paper, these objects are denoted as \emph{Object 1}, \emph{Object 2}, and \emph{Object 3}. They are shown in Fig.~\ref{fig:all-obj}. Although the objects have comparable size, they exhibit different appearance, texture, and local geometry. This makes them suitable for evaluating whether the learned visual corrections remain object specific while still being robust to differences in contour and visual structure. Separate datasets were generated for each object, and the same object set was later reused in the real world manipulation experiments.

The generated datasets were designed to expose the model to progressively harder pose correction problems. The main factor used to define dataset complexity is the number of degrees of freedom (DOF) that vary between the current pose $P_i$ and the target pose $P_0$. Four trajectory families were considered:
\begin{itemize}
    \item \textbf{2-DOF (Planar):} the robot follows a zig-zag trajectory on the $x$-$y$ plane within a square region of side length $l$, while keeping the height $z$ and the camera orientation fixed.
    \item \textbf{3-DOF (Planar with rotation):} the planar trajectory is preserved, but an additional in-plane rotation around the $z$ axis is introduced.
    \item \textbf{3-DOF (Cone):} a grid of planar positions is combined with vertical displacement in $z$, and the robot executes trajectories between those positions and $P_0$, producing a coarse approach dataset.
    \item \textbf{4-DOF (Cone with rotation):} the cone configuration is extended by also varying the in-plane rotation around the $z$ axis.
\end{itemize}


The number of samples obtained for each object and each trajectory family is reported in Table~\ref{tab:dataset_size}. The larger number of samples in the cone based datasets is consistent with their longer trajectories and larger traveled distance, which increase the number of captured frames during data collection.

\begin{table}[t]
\centering
\caption{Number of samples in each automatically generated dataset for each object.}
\label{tab:dataset_size}
\resizebox{\columnwidth}{!}{%
\begin{tabular}{lcccc}
\hline
\textbf{Object} & \textbf{2-DOF} & \textbf{3-DOF} & \textbf{3-DOF} & \textbf{4-DOF} \\
 & \textbf{Planar} & \textbf{Planar + rot.} & \textbf{Cone} & \textbf{Cone + rot.} \\
\hline
Object 1 & 1485 & 1948 & 2717 & 2842 \\
Object 2 & 1119 & 2090 & 2723 & 2831 \\
Object 3 & 1121 & 2117 & 2725 & 2785 \\
\hline
\end{tabular}%
}
\end{table}



\subsection{Visual Policy Learning}

The visual policy is trained to regress the relative pose correction required to move the robot from the current pose to the target pose. The adopted model follows the architecture proposed by Johns \cite{johns2021}. It consists of four convolutional layers with max pooling and ReLU activation, followed by four fully connected layers that regress translation and rotation from a single RGB image. Each training sample contains an image and a seven dimensional target vector composed of three translation values and a quaternion with four rotation components.

The training process is implemented in PyTorch. Data loading, normalization, metric computation, and inference routines are defined in Python modules connected to the ROS-side application. The loss function is defined as
\begin{equation}
\mathcal{L} = \mathrm{MSE}_{\text{trans}} + 0.01 \cdot \mathrm{MSE}_{\text{rot}}
\end{equation}
where $\mathrm{MSE}_{\text{trans}}$ is the mean squared error for translation and $\mathrm{MSE}_{\text{rot}}$ is the mean squared error for rotation. The weighting factor of $0.01$ gives greater importance to translation accuracy, which is the most relevant term for the approach and grasp alignment stages.

An adaptive learning rate is used during training. The initial value is set to $10^{-4}$ and reduced by half when the validation loss does not improve sufficiently over a predefined patience interval. Early stopping is also employed to terminate training when further progress is no longer observed. Once training is complete, the learned weights are stored in the corresponding dataset directory and later loaded by the inference node during execution. 

\vspace{-5mm}

\section{Results}

\subsection{Simulation Results}

Simulation experiments were conducted in Isaac Sim to evaluate the closed loop behavior of the proposed coarse to fine controller before deployment on the physical robot. In this setup, a synthetic dataset was generated in a simple scene with a blue background and a black cube, and the object remained fixed during testing. The robot was initialized from nearby poses, in a procedure analogous to the one used during dataset generation.

To evaluate convergence under different initial conditions, nine starting positions were generated automatically around the target pose. 
From each start point, the robot first executed the coarse approach stage using the model trained on the cone dataset. The resulting trajectories are shown as solid colored lines. This stage was terminated when the predicted vertical correction became smaller than 8~mm or when the robot height reached the safety threshold of 160~mm with respect to the base.


\begin{table*}[t]
\centering
\caption{Evaluation-set error for each object and each dataset family. Translation is reported in meters and rotation in radians.}
\label{tab:eval_error}
\begin{tabular}{lcccccccc}
\hline
\multirow{2}{*}{\textbf{Object}} & \multicolumn{2}{c}{\textbf{2-DOF}} & \multicolumn{2}{c}{\textbf{3-DOF}} & \multicolumn{2}{c}{\textbf{3-DOF}} & \multicolumn{2}{c}{\textbf{4-DOF}} \\
 & \multicolumn{2}{c}{\textbf{Planar}} & \multicolumn{2}{c}{\textbf{Planar + rot.}} & \multicolumn{2}{c}{\textbf{Cone}} & \multicolumn{2}{c}{\textbf{Cone + rot.}} \\
 & \textbf{m} & \textbf{rad} & \textbf{m} & \textbf{rad} & \textbf{m} & \textbf{rad} & \textbf{m} & \textbf{rad} \\
\hline
Object 1 & $6.8\times10^{-3}$ & $8.2\times10^{-3}$ & $7.5\times10^{-3}$ & 0.18  & $5.3\times10^{-3}$ & $2.5\times10^{-4}$ & $8.1\times10^{-3}$ & 0.10 \\
Object 2 & $7.5\times10^{-3}$ & $2.9\times10^{-4}$ & $8.2\times10^{-3}$ & 0.889 & $7.8\times10^{-3}$ & $3.2\times10^{-4}$ & $6.2\times10^{-3}$ & 0.09 \\
Object 3 & $9.8\times10^{-3}$ & $4.1\times10^{-4}$ & $8.6\times10^{-3}$ & 0.21  & $8.3\times10^{-3}$ & $3.5\times10^{-4}$ & $1.0\times10^{-2}$ & 0.11 \\
\hline
\end{tabular}
\end{table*}

\begin{figure}[b]
    \centering
    \begin{subfigure}{0.2\textwidth}
        \includegraphics[width=\linewidth]{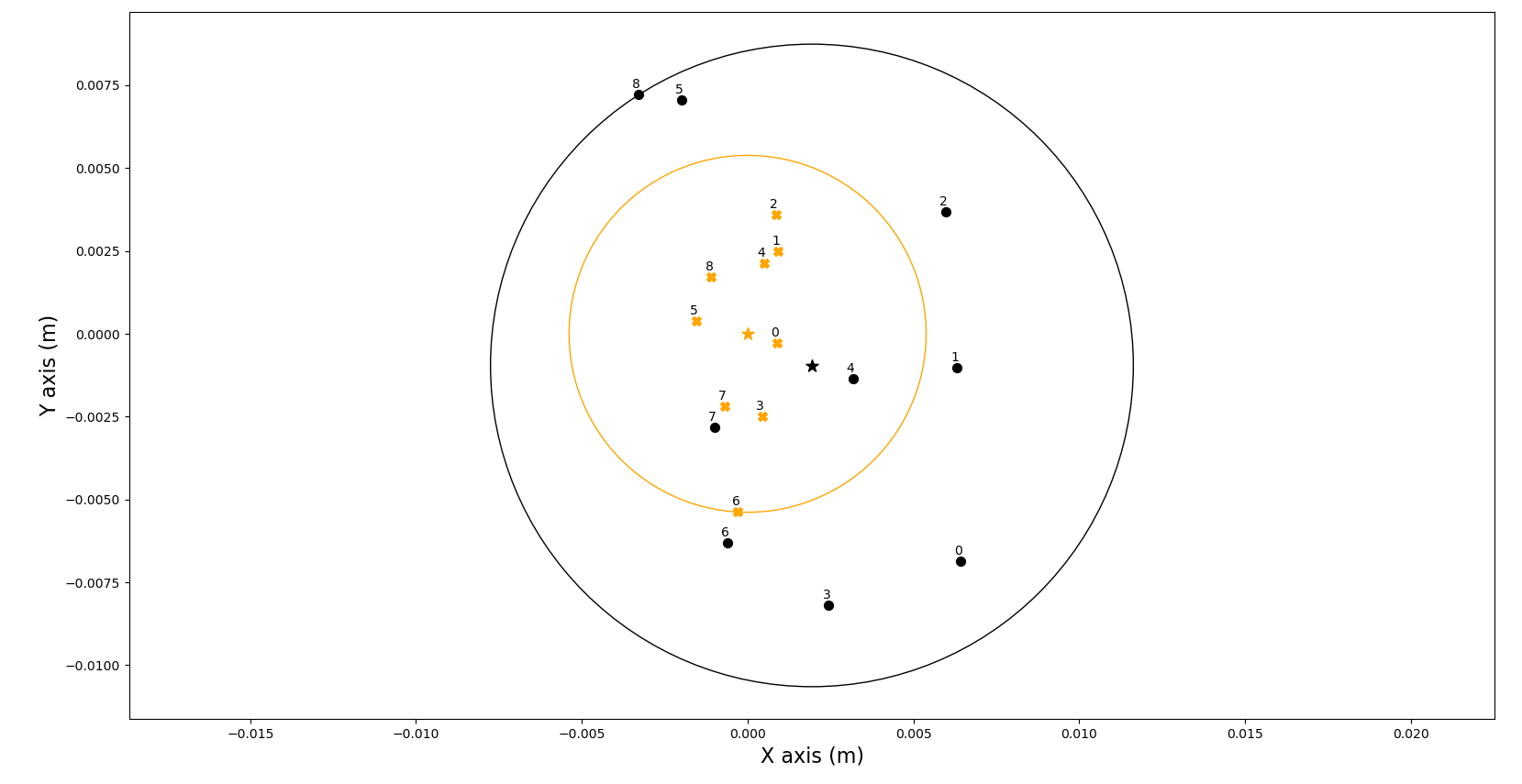}
        \caption{XY plane}
        \label{fig:plane-final-dots}
    \end{subfigure}
    \hfill
    \begin{subfigure}{0.2\textwidth}
        \includegraphics[width=\linewidth]{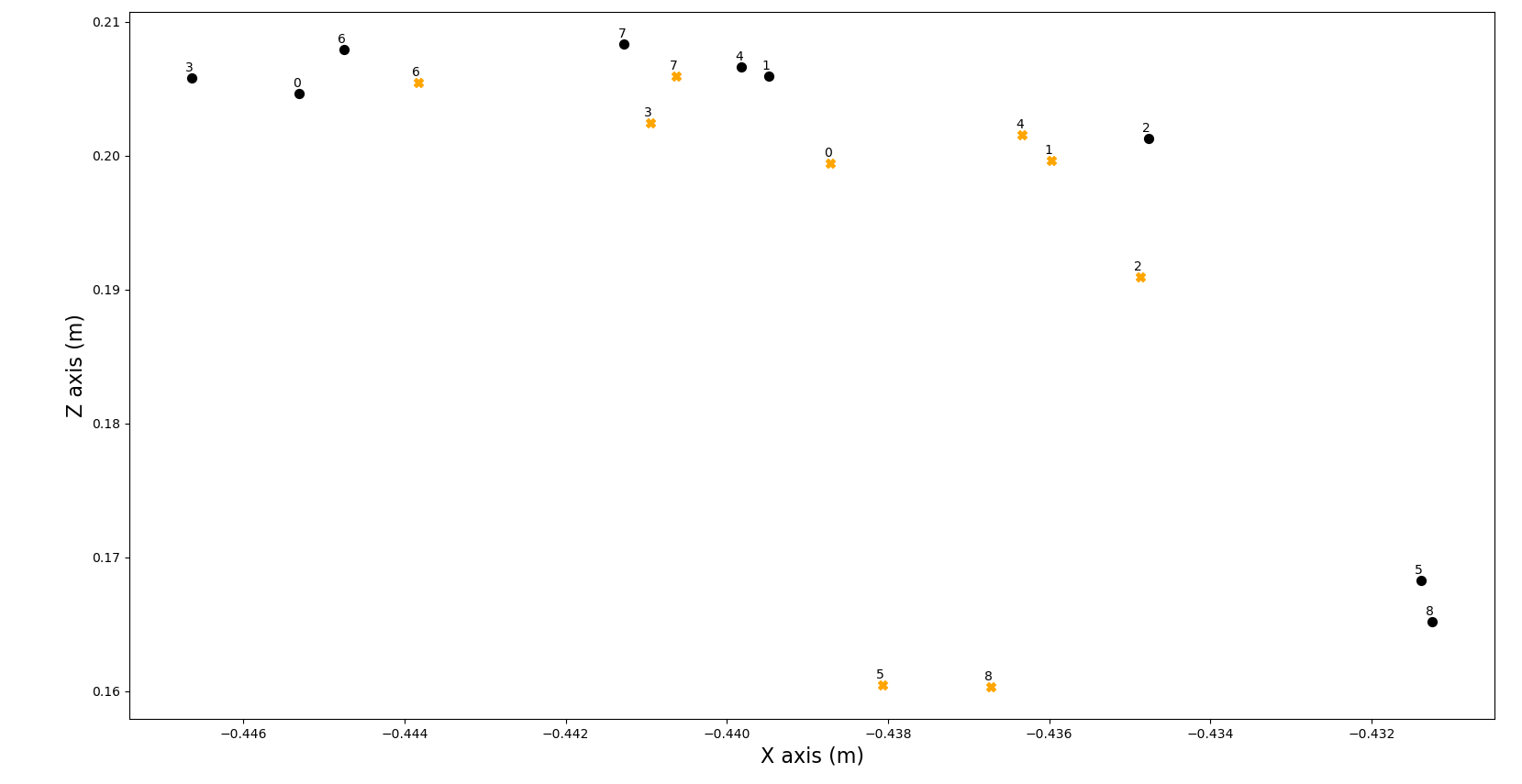}
        \caption{XZ plane}
        \label{fig:height-final-dots}
    \end{subfigure}
    \caption{Final positions of the simulated trajectories after coarse approach and planar refinement.}
    \label{fig:final-points}
\end{figure}

The final positions reached after the approach and refinement stages are summarized in Fig.~\ref{fig:final-points}. In Fig.~\ref{fig:plane-final-dots}, we show the projection on the XY plane, the black points correspond to the end of the coarse stage and the orange crosses correspond to the end of the refinement stage. The centroids and enclosing circles show that the planar spread was reduced after refinement. In particular, the radius of the enclosing circle decreased from 9.69~mm for the coarse stage to 5.38~mm for the refinement stage, indicating that the second stage effectively improved the final lateral convergence.

In Fig.~\ref{fig:height-final-dots}, which shows the projection on the XZ plane, a different behavior can be observed. Two trajectories, corresponding to cases 5 and 8, ended at noticeably lower heights than the others, indicating less stable depth estimation. The average final height after the coarse stage was 197~mm, and increased to 206~mm if those two outliers were excluded. The average vertical difference between the end of the coarse stage and the end of refinement was 5.3~mm when all trajectories were considered, and 5.1~mm when the outliers were excluded. In addition, the radius of the enclosing three dimensional sphere only decreased from 33.39~mm to 31.55~mm. 

Overall, the simulation results support the proposed coarse to fine strategy. The approach model was able to drive the robot toward the target region from multiple initial positions, while the planar refinement model consistently reduced the lateral dispersion of the final estimates. At the same time, the simulation experiments also revealed that depth prediction remained more challenging than planar correction, which anticipates some of the limitations later observed in the real world evaluation.

\subsection{Real World Evaluation}

The learned models were deployed on a physical UR5e robot equipped with an RG6 gripper and a wrist mounted monocular camera. The same three objects used in dataset generation were used during testing, denoted here as \emph{Object 1}, \emph{Object 2}, and \emph{Object 3}. Each trial started from a pose approximately 12 cm above the reference pose used during data generation, and the object was placed manually within the camera field of view. The controller then executed the same three-stage procedure used in simulation: coarse approach, planar refinement, and grasp execution. Experiments were conducted on a representative subset of objects with varying appearance and geometry to evaluate robustness under different visual conditions. 

To complement the end to end grasp trials, the trained models were evaluated on a held out subset. Following the evaluation protocol adopted in the thesis, translation and rotation errors were reported separately. Translation error corresponds to the Cartesian distance between the predicted and ground truth positions, while rotation error corresponds to the sum of angular differences across the three rotation axes. Table~\ref{tab:eval_error} summarizes these values for each object and each dataset family. In the analysis, a prediction may be considered correct when the translation error is below 5 mm and the rotation error is below $5^\circ$ (0.087 rad).


The grasp success rates under both conditions are summarized in Table~\ref{tab:success_combined}. Without object rotation, Object~1 and Object~2 achieved the highest performance, with success rates of 66.6\% and 63.6\%, respectively, while Object~3 obtained a lower success rate. When object rotation was introduced, performance decreased overall, with Object~1 dropping to 36.4\%, Object~2 maintaining the best performance at 55.5\%, and Object~3 decreasing to 25.0\%. The end to end grasp experiments provide the clearest measure of practical manipulation performance. First, trials were conducted without rotating the object with respect to the camera frame. The experiments were then repeated with additional in-plane rotation around the camera $z$ axis. The grasp success rates under both conditions are summarized in Table~\ref{tab:success_combined}. Without object rotation, Object~1 and Object~2 achieved the highest performance, with success rates of 66.6\% and 63.6\%, respectively, while Object~3 obtained a lower success rate. When object rotation was introduced, performance decreased overall, with Object~1 dropping to 36.4\%, Object~2 maintaining the best performance at 55.5\%, and Object~3 decreasing to 25.0\%.

\begin{table}[t]
\centering
\caption{Grasp success rate in real-world experiments under different conditions.}
\label{tab:success_combined}
\begin{tabular}{lcc}
\hline
\textbf{Object} & \textbf{No rotation (\%)} & \textbf{With rotation (\%)} \\
\hline
Object 1 & 66.6 & 36.4 \\
Object 2 & 63.6 & 55.5 \\
Object 3 & 35.2 & 25.0 \\
\hline
\end{tabular}
\end{table}


Taken together, the real-world results show three main points. First, the proposed self-supervised pipeline is not restricted to simulation and can support closed-loop grasp attempts on a physical UR5e platform. Second, the performance is object dependent, suggesting sensitivity to appearance, geometry, and orientation. Third, the success rate degradation under rotation confirms that visual alignment becomes more difficult when the observed image departs from the nominal demonstration distribution. Even so, all tested objects achieved successful grasp attempts under at least one experimental condition, which supports the practical viability of the proposed approach.

Although no direct baseline comparison is included, the proposed method differs from classical visual servoing approaches by learning pose corrections directly from RGB observations without explicit feature extraction or calibration. Compared to demonstration-based methods requiring human input, the proposed approach removes manual supervision through automatic data generation.

\vspace{-5mm}
\section{Conclusion}

This paper presented a self-supervised framework for visual robotic manipulation based on automatically generated demonstrations. The robot collected image-pose pairs around a reference manipulation pose and learned relative pose corrections from single-frame RGB observations. A coarse approach model and a planar refinement model were combined in a closed-loop controller that transferred from simulation to a real UR5e platform with minimal interface changes through ROS~2.

The results showed improved planar convergence in simulation and successful real-world grasp attempts on multiple objects with varying appearance and geometry, without explicit camera extrinsic calibration. At the same time, the method remained sensitive to depth estimation, object geometry, and object rotation. Future work should therefore focus on domain randomization, richer sensing such as RGB-D or stereo, explicit object localization, and tighter integration of visual inference with kinematic constraints.

\vspace{-10mm}

\section{Acknowledgements}

The authors of this work would like to thank the Technological University of Uruguay and the Laboratory of Robotics and AI for the support in this work.

\vspace{-5mm}

\bibliographystyle{./bibliography/IEEEtran}
\bibliography{./bibliography/IEEEabrv,./bibliography/main}

\end{document}